%% file: root.tex
\documentclass[letterpaper, 10 pt, conference]{ieeeconf}

\IEEEoverridecommandlockouts
\usepackage{multicol}
\usepackage[bookmarks=true]{hyperref}
\newif\ifdraft
\draftfalse   

\ifdraft
  \usepackage[colorinlistoftodos,prependcaption,textsize=small]{todonotes}
\else
  \usepackage[disable]{todonotes}
\fi

\usepackage{amsmath} 
\usepackage{caption}
\usepackage{booktabs}
\usepackage{times}
\usepackage{amssymb} 
\usepackage{amsfonts}
\usepackage{biblatex}
\usepackage{balance}

\begin{document}

\title{\LARGE \bf
Rapid Embodiment Adaptation for Quadrupedal Locomotion
}

\author{
  Dichen Li$^{1}$,
  Bo Ai$^{1,2}$,
  Nico Bohlinger$^{3}$,
  Jan Peters$^{3,4,6}$,
  Hao Su$^{5}$,
  Henrik I. Christensen$^{1}$ \vspace{5pt}
  \\ 
    $^1$UC San Diego \quad 
    $^2$Stanford University \quad 
    $^3$TU Darmstadt \quad 
    $^4$hessian.AI \quad
    $^5$Sudo AI GmbH \\
    $^6$German Research Center for AI (DFKI) \& Robotics Institute Germany \\ 
    \authorblockA{\textbf{\textcolor{magenta}{\url{https://embodiment-adaptation.github.io/}}}} \vspace{-15pt}
\thanks{
Funded by the NCN (UMO-2021/43/I/ST6/02711), DFG (PE 2315/17-1).
\newline Corresponding author: Dichen Li $\langle$\tt\small dil012@ucsd.edu$\rangle$
}}

\maketitle
\setlength{\abovedisplayskip}{1pt}
\setlength{\belowdisplayskip}{1pt}

\thispagestyle{empty}
\pagestyle{empty}

\input{text/macros}

\input{text/000_abstract}

\input{text/010_intro}
\input{text/020_related-works}
\input{text/030_method}

\input{text/050_results}
\input{text/060_conclusion}


\balance
\printbibliography

\input{text/070_appendix}

\end{document}

%% file: text/macros.tex
\newcommand{\secref}[1]{Section~\ref{#1}}
\renewcommand{\eqref}[1]{Eqn~\ref{#1}}
\newcommand{\figref}[1]{Figure~\ref{#1}}
\newcommand{\subfig}[1]{\textit{#1}}
\newcommand{\tabref}[1]{Table~\ref{#1}}
\newcommand{\taskref}[1]{Task~\ref{#1}}
\newcommand{\chapref}[1]{Chapter~\ref{#1}}
\newcommand{\appendref}[1]{Appendix~\ref{#1}}
\newcommand{\eqnref}[1]{\eqref{#1}}
\newcommand{\algoref}[1]{Algorithm~\ref{#1}}
\newcommand{\subfigref}[2]{\figref{#1}(\subfig{#2})}
\newcommand{\envref}[1]{Env~\ref{#1}}
\newcommand{\theoref}[1]{Theorem~\ref{#1}}

\newcommand{\ie}{\textrm{i.e.}}
\newcommand{\eg}{\textrm{e.g.}}
\newcommand{\etc}{\textrm{etc.}}
\newcommand{\etal}{\textrm{et~al.}}

%
%
\def\appro{\textasciitilde}

%
%
\newcommand\norm[1]{\left\lVert#1\right\rVert}
\def\adist{$\mathcal{A}$-distance}
\def\X{\mathcal{X}}
\def\Y{\mathcal{Y}}
\def\Z{\mathcal{Z}}
\def\D{\mathcal{D}}
\def\S{\mathcal{S}}
\def\A{\mathcal{A}}
\def\H{\mathcal{H}}
\def\T{\mathcal{T}}
\def\T{\mathcal{J}}
\def\T{\mathcal{T}}
\def\L{\mathcal{L}}
\def\J{\mathcal{J}}
\def\M{\mathcal{M}}
\def\N{\mathcal{N}}
\def\O{\mathcal{O}}
\def\R{\mathcal{R}}

\newcommand{\rebuttal}[1]{\textcolor{black}{#1}}

%
%
\def\method{Method\xspace}

%
%
\def\MSE{MSE}
\def\EMD{EMD}
\def\pushes{\# Pushes}
\def\SR{Success Rate}

%
%
\def\weblink{\urllink[pre = \bgroup\bf, post = \egroup]}
\newcommand{\cell}[1]{\#{#1}}
\newcommand{\highlight}[1]{\textbf{\textit{#1}}}
\newcommand{\bighighlight}[1]{\textbf{\textit{#1}}}
\newcommand{\pinklink}[2]{\weblink{#1}{\textcolor{magenta}{#2}}}

\newenvironment{customquote}%
  {\begin{mdframed}[topline=false, bottomline=false, rightline=false,%
    linewidth=1pt,linecolor=black,%
    leftmargin=1cm,rightmargin=1cm,%
    innerleftmargin=10pt,innerrightmargin=10pt,%
    innertopmargin=0pt,innerbottommargin=0pt,%
    skipabove=\baselineskip,skipbelow=\baselineskip]%
   \fontfamily{ppl}\selectfont
  }%
  {\end{mdframed}}

\def\reviewerone{\textcolor{purple}{t1pD}}
\def\reviewertwo{\textcolor{Dandelion}{YYPj}}
\def\reviewerthree{\textcolor{cyan}{6Tyi}}

\newcommand\blfootnote[1]{%
\begingroup 
\renewcommand\thefootnote{}\footnote{#1}%
\addtocounter{footnote}{-1}%
\endgroup 
}

\newcommand{\bo}[1]{\textcolor{purple}{[Bo: #1]}}
\newcommand{\nico}[1]{\textcolor{blue}{[Nico: #1]}}
\newcommand{\dichen}[1]{\textcolor{orange}{[Dichen: #1]}}

\def\pushingtask{\textit{Object Pushing}}
\def\reshapingtask{\textit{Plasticine Reshaping}}

\def\qone{\textbf{Q1}}
\def\qtwo{\textbf{Q2}}
\def\qthree{\textbf{Q3}}

%% file: text/000_abstract.tex
\begin{abstract}

Humans readily adapt their movements as their bodies change through aging, injury, or load carrying, but learning-based robot policies often break when hardware properties shift. We introduce an online embodiment adaptation framework for quadrupedal locomotion that infers embodiment parameters from short interaction histories and conditions control on the inferred hardware state. Our method pairs a generalist policy trained under embodiment randomization with a lightweight adaptation module that identifies physical changes within half a second. We evaluate two representative forms of embodiment variation: joint-range constraints and trunk-mass changes, corresponding to joint-level kinematic degradation and body-level dynamic variation. In simulation, the module accurately estimates these changes and enables closed-loop control that substantially outperforms policies conditioned directly on interaction history. On a real Unitree Go2 robot, our system maintains stable locomotion under severe instances of the evaluated changes, including a fully locked leg and a 5 kg payload, where non-adaptive methods fail. These results demonstrate the practicality of explicit online embodiment identification for rapid adaptation to joint-limit and payload-mass changes, and provide a step toward handling broader forms of uncertain, degraded, or changing robot hardware.

\end{abstract}

%% file: text/010_intro.tex
\section{Introduction}

Cross-embodiment learning aims to build generalist policies that can operate across diverse robot platforms. Recent progress has demonstrated shared controllers across morphologies in locomotion~\cite{ai2025towards, bohlinger2025multi, bohlinger2024onepolicy, liu2025locoformer} and robot manipulation~\cite{he2025scaling, patel2024getzero, wei2024dro, intelligence2026pi}. However, most prior work assumes a known and static embodiment at deployment time, including information on joint limits, link masses, inertial properties, and the kinematic structure. In contrast, humans continually recalibrate to embodiment changes such as injury, aging, fatigue, load carrying, and tool use~\cite{Nezafat2001Long-term, winter1990biomechanics, Maravita2004Tools}. This adaptivity suggests that the robot embodiment should be treated not as a static input, but as a variable that is inferred online for control. This raises a fundamental question: how can a cross-embodiment policy identify and adapt to unknown or changing embodiment parameters during deployment?


System identification provides a principled framework for estimating physical parameters. However, traditional methods that are formulated as an optimization problem require collecting calibration datasets in the target domain. In robot learning, prior work has focused primarily on estimating object properties~\cite{ai2024robopack} and environment parameters~\cite{kumar2021rma, zhang2025alearning}; the online identification of the robot's internal embodiment, such as joint range constraints, remains underexplored. Recent learning-based methods address this by mapping interaction histories directly to actions end-to-end~\cite{liu2025locoformer}, but their implicit adaptation requires long histories (\eg, several seconds of observations), incurring substantial training overhead and slow deployment-time adaptation.



\begin{figure*}[t]
    \centering
    \includegraphics[width=1.0\textwidth]{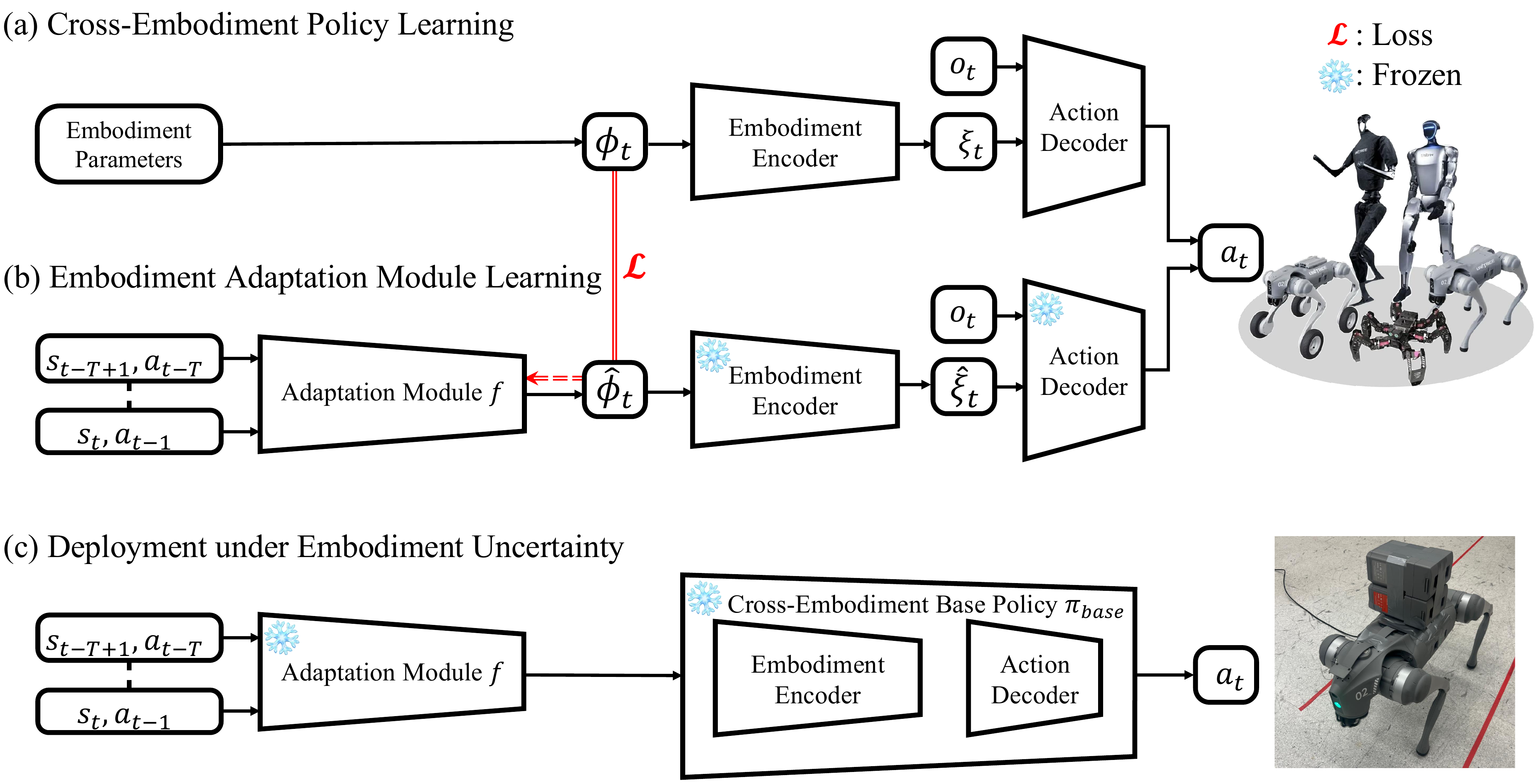} 
    \captionof{figure}{
        \textbf{Pipeline of the proposed online embodiment adaptation framework. }
        The cross-embodiment base policy $\pi_{\text{base}}$ is trained to handle locomotion tasks across different embodiments with the ground-truth embodiment description. The adaptation module $f$ is trained to infer embodiment parameters from observation history for the base policy to condition on. The above adaptation method (Explicit Representation) infers physical embodiment parameters $\hat{\phi}_t$. An alternative method (Latent Representation) for inferring latent embodiment representations $\hat{\xi}_t$, not shown here, is described in \ref{sec:adaptation_module}.
        \vspace{-15pt}
    }
    \label{fig:dual_loop_pipeline}
\end{figure*}

To enable efficient training and rapid adaptation, we decompose control into two separately trained modules (\figref{fig:dual_loop_pipeline}): (i) an embodiment adaptation module and (ii) a cross-embodiment policy. The adaptation module infers embodiment parameters online from short-horizon action-state histories, estimating quantities such as joint-limit scaling and trunk mass offset. These parameters represent two complementary forms of embodiment variation: joint-level kinematic constraints and robot-level dynamic changes. We study these variations on a fixed nominal morphology as a focused setting for online embodiment identification. The cross-embodiment policy, trained across randomized variations of these parameters, is explicitly conditioned on the inferred embodiment and generates control commands aligned with the current hardware configuration. This two-stage training provides direct supervision for each module, enabling efficient training and rapid parameter identification at inference compared with implicit end-to-end alternatives.


We validate the framework in both simulation and the real world. In simulation, the adaptation module identifies joint-limit scale and trunk mass offset within only 0.4 seconds of interaction, enabling closed-loop control that substantially outperforms the end-to-end implicit adaptation baseline across the evaluated settings. In the real world, the Unitree Go2 maintains stable locomotion under severe joint-range constraints and payload changes, including a heavily constrained front-right leg and a 5.0 kg payload, where policies without online adaptation fail. These results demonstrate rapid and robust adaptation to the evaluated joint-limit and mass variations, providing a step toward cross-embodiment policies that can handle broader forms of unknown or changing robot parameters.

%% file: text/020_related-works.tex
\section{Related Work} 
\label{sec:related_work}

\subsection{Cross-Embodiment Learning}

Cross-embodiment learning seeks to build generalist policies that learn from and generalize across robots of varying morphologies. Progress has been visible in manipulation, where controllers can operate across multi-fingered hands~\cite{he2025scaling, patel2024getzero, wei2024dro} or broader form factors including robot arms, grippers, and mobile manipulators~\cite{intelligence2026pi}. In legged locomotion, most policies remain specific to a single platform~\cite{rudin2022learning, zhuang2023, cheng2023, bohlinger2025gait, stasica2025bridge, cheng2024expressive}, motivating a line of work on universal locomotion controllers.

A central challenge is that different robots have variable observation and action spaces, varying joint counts, kinematic topologies, and actuator properties. Early approaches captured morphological structure with Graph Neural Networks (GNNs), which rely on morphology-specific graph designs and struggle to generalize across distinct morphology families~\cite{wang2018, huang2020one}. Transformer-based methods improved on this by globally aggregating embodiment information, though they often depend on manually crafted tokens or fixed joint-type vocabularies that limit transfer to unseen morphologies~\cite{gupta2022, sferrazza2025body, patel2025get}. The Unified Robot Morphology Architecture (URMA)~\cite{bohlinger2024onepolicy} took a further step by explicitly conditioning the policy on per-joint and general robot parameters; subsequent work has scaled this paradigm with stronger embodiment randomization~\cite{bohlinger2025multi} and shown that increasing training-embodiment diversity improves generalization~\cite{ai2025towards}.

These methods, however,  assume that an accurate embodiment description is available at deployment, an assumption that breaks under wear-and-tear, changing payloads, or active modifications to the robot's body~\cite{nygaard2021real}. Our work targets this gap by enabling an embodiment-conditioned controller to adapt online when robot-side parameters are unknown or changing.

\subsection{Online Embodiment Identification}

Online system identification estimates physical context from interaction during deployment. In robot learning, prior work has inferred object properties from interaction histories~\cite{ai2025review, ai2024robopack, xu2019densephysnet, qi2022hand}, as well as environmental and dynamic factors such as terrain geometry, friction, payload, and actuator properties~\cite{kumar2021rma, kumar2022adapting, miki2022learning, choi2023learning, valsecchi2024accurate}. RMA-style methods~\cite{kumar2021rma, kumar2022adapting}, in particular, train policies with privileged physical context and learn adaptation modules that recover this context from recent interaction history. Our framework follows this paradigm but applies it to a structured embodiment description that includes both body-level dynamics, such as payload, and per-joint kinematic properties, such as joint ranges and locking. While adaptation to dynamic variation has been studied, online identification of robot-side kinematic changes remains comparatively underexplored. Related implicit approaches instead map interaction histories directly to actions, allowing task-relevant embodiment information to emerge without explicit parameter estimation~\cite{liu2025locoformer}. However, they typically require histories spanning several seconds. We use direct supervision to infer an explicit or latent embodiment representation from a short interaction history and condition a cross-embodiment policy on the estimate, enabling rapid deployment-time adaptation.



%% file: text/030_method.tex
\section{Method}
\subsection{Overview}

We consider cross-embodiment learning for robotics, where control policies must generalize across diverse and potentially changing embodiments. We treat the embodiment parameters $\phi_t$, such as joint limits, torque limits, and mass properties, as hidden context variables that affect the robot dynamics and the optimal control strategy.

At time $t$, the robot receives an observation $o_t \in \mathcal{O}$, consisting of proprioceptive state and command inputs, and executes an action $a_t \in \mathcal{A}$. The system then transitions according to dynamics that depend on both the physical state and the current embodiment parameters $\phi_t$. Since $\phi_t$ is not directly observed at deployment, the policy cannot condition on the true embodiment parameters and must infer them from interaction.

Let
\[
h_t = \{(o_{t-T+1}, a_{t-T}), \dots, (o_t, a_{t-1})\}
\]
denote a short sliding window of past observations and actions. We introduce an adaptation module $f$ that estimates the current embodiment context from this history,
\[
\hat{\phi}_t = f(h_t),
\]
and a cross-embodiment policy $\pi$ that predicts actions conditioned on the current observation and the inferred embodiment:
\[
a_t \sim \pi(\cdot \mid o_t, \hat{\phi}_t).
\]
This forms a closed-loop adaptation process: as the robot interacts with the environment, the history $h_t$ is updated, the embodiment estimate $\hat{\phi}_t$ is refined online, and the policy adjusts its actions accordingly.

Our objective is to maximize the expected return under the true embodiment:
\[
\max_{\pi,f}\; \mathbb{E}\Big[\sum_{t=0}^{H} \gamma^t\, R(s_t,a_t;\phi_t)\Big],
\]
where the reward and transition dynamics depend on the unobserved embodiment parameters $\phi_t$. In practice, we train $\pi$ with reinforcement learning using oracle embodiment information in simulation, and train $f$ via supervised learning to recover the embodiment context from interaction histories.

Next, we introduce the cross-embodiment policy learning (\secref{sec:policy_learning}), and the adaptation module (\secref{sec:adaptation_module}). 

\subsection{Cross-Embodiment Policy} \label{sec:policy_learning}
We use URMA \cite{bohlinger2024onepolicy} as the architecture for our cross-embodiment base policy, which we denote as \(\pi_{\text{base}}(a_t \mid o_t, \phi)\).
URMA consists of a latent encoder $E_{\text{lat}}$ and an action decoder $D_{\text{act}}$.
Similar to \cite{bohlinger2025multi}, we train \(\pi_{\text{base}}\) with deep reinforcement learning with diverse online embodiment randomization, including joint limits and mass distributions of the robot.
Once trained, the parameters of the URMA policy are frozen for all subsequent online adaptation experiments.

\subsection{Embodiment Adaptation Module} \label{sec:adaptation_module}

The goal of the module is to estimate the embodiment paramters from interaction history $h_t$. Following prior morphology-conditioned policy design~\cite{bohlinger2024onepolicy,ai2025towards}, the parameters include joint-level (\eg, joint limit) and body-level information (\eg, body mass). 



\subsubsection{Embodiment representation} \label{sec:parameterization}
The adaptation module can predict either physical embodiment parameters  $\phi_t$ or learned latent representations $\xi_t$ that emerge from cross-embodiment policy training. We investigate both representations in this work.


\textbf{Explicit representation.}
 $f_{\text{exp}}$ predicts interpretable physical parameters consisting of per-joint parameters $\hat{d}_t$ and a global trunk mass term $\hat{m}_t$:
\[
(\hat{d}_t, \hat{m}_t) = f_{\text{exp}}(h_t), \quad
\hat{d}_{t,j}=(\hat{q}^0_{t,j}, \hat{\tau}^{\max}_{t,j}, \hat{q}^{\min}_{t,j}, \hat{q}^{\max}_{t,j}),
\]
where $\hat{q}^0_{t,j}$ is the nominal joint position, $\hat{\tau}^{\max}_{t,j}$ is the torque limit, and $\hat{q}^{\min}_{t,j}, \hat{q}^{\max}_{t,j}$ are the joint range endpoints.
We construct the full explicit embodiment description $\hat{\phi}_t$ by combining the predicted adaptable parameters with fixed structural descriptors $d^{\text{fix}}$, for all joints $J$:
\[
\hat{\phi}_{t,j} = (d^{\text{fix}}_j, \hat{d}_{t,j}), \qquad
\hat{\phi}_t = \Big[\{\hat{\phi}_{t,j}\}_{j=1}^{J}, \hat{m}_t\Big],
\]
The control action is then produced by the frozen base policy:
\(
a_t \sim \pi_{\text{base}}(\cdot \mid o_t, \hat{\phi}_t).
\)

\textbf{Latent representation.}
$f_{\text{lat}}$ predicts latent embodiment embeddings $\hat{z}_t$ of size $d_z$, where $d_z$ is a hyperparameter:
\[
(\hat{z}_t, \hat{m}_t) = f_{\text{lat}}(h_t), \qquad \hat{z}_t \in \mathbb{R}^{J \times d_z},
\]
with $z_t = E_{\text{lat}}(d_t)$ denoting the latent embedding of ground-truth explicit adaptable parameters $d_t$. This is lower-dimensional and potentially more structured~\cite{ai2025towards}, since the latent representation is optimized to capture the most useful information for control.
We construct the full latent embodiment representation $\hat{\xi}_t$:
\[
\hat{\xi}_t = \big[d^{\text{fix}}, \hat{z}_t, \hat{m}_t\big].
\]
Given the latent representation $\hat{\xi}_t$, the latent encoder $E_{\text{lat}}$ of the base policy is skipped, and $\hat{\xi}_t$ is directly input to the frozen action decoder $D_{\text{act}}$ of the base policy to produce the action:
\(
a_t \sim D_{\text{act}}(\cdot \mid o_t, \hat{\xi}_t).
\)

\subsubsection{Network architecture}
The adaptation module predicts explicit or latent embodiment representations using the same model architecture. It first uses a Long Short-Term Memory (LSTM)~\cite{schmidhuber1997long} encoder to capture temporal dependencies from the interaction history. The encoded features are then processed by a Transformer backbone that performs attention across the joint dimension, enabling joint-wise aggregation of information. The network predicts trunk mass $\hat{m}_t$ and joint-level information $\hat{d}_t$ or $\hat{z}_t$ separately. 

\subsubsection{Training}
We train the adaptation module $f$ on offline trajectories collected in simulation under randomized embodiment variations.
For each trajectory, we record the ground-truth adaptable embodiment parameters, consisting of per-joint parameters $d_t$ and trunk mass $m_t$.
Both variants minimize the same supervised learning objective,
\begin{equation}
    \label{eqn:adapt_loss}
    \min_{f}\; \mathbb{E}_{t}\Big[\big\|f(h_t) - \langle \rho_t, m_t \rangle \big\|^2\Big],
\end{equation}
where $f=f_{\text{exp}}$ and $\rho_t=\hat{\phi}$ for explicit embodiment representation, and $f=f_{\text{lat}}$ and $\rho_t=\hat{\xi}_t $ for latent embodiment representation.




%% file: text/050_results.tex
\section{Experimental Setup} \label{sec:exp_setup}


    
\subsection{Task Setup}
We evaluate the proposed framework on the standard command-velocity-tracking locomotion task~\cite{bohlinger2024onepolicy, bohlinger2025multi, ai2025towards, kumar2021rma}, testing its robustness to changes in embodiment parameters during an episode. In simulation, we sample longitudinal, lateral, and yaw-velocity commands, covering multidirectional locomotion. For the real-world demonstration, we use a fixed forward-velocity command on flat ground to enable controlled and clean comparisons.

We focus on two types of embodiment changes: joint-limit variations and trunk-mass variations. Joint-limit variations simulate actuator degradation or partial joint locking by scaling or restricting the range of motion. Trunk-mass offsets simulate payload additions, resulting in changes in body dynamics. These modifications represent joint-level kinematic constraints and body-level dynamic changes, respectively. We refer to them as joint-limit scaling, which multiplicatively scales the original joint range, and trunk-mass offset, which additively modifies the trunk mass.

\textbf{Hardware setup.}
We use the Unitree Go2 quadruped as the robot platform. The robot's nominal mass is 15.7 kg, including the battery, and it has 12 actuated degrees of freedom, with three joints per leg (hip, thigh, and knee). During deployment, we impose joint constraints by applying a high control gain outside a prescribed joint range and introduce mass changes by manually adding payloads..



\begin{table}[t]
\centering
\caption{\textbf{Reward terms in RL policy training.}}
\label{tab:rl_reward}
\begin{tabular}{lc}
\toprule
\textbf{Reward Term} & \textbf{Coefficient} \\
\midrule
Tracking XY Velocity Command & $2.0 \cdot \Delta t$ \\
Tracking Yaw Velocity Command & $1.0 \cdot \Delta t$ \\
Pitch Roll Position & -$5.0 \cdot \Delta t$ \\
Base Height & -$30.0 \cdot \Delta t$ \\
Base Z Velocity & -$2.0 \cdot \Delta t$ \\
Air Time & $0.2 \cdot \Delta t$ \\
Stand Still & -$5.0 \cdot \Delta t$ \\
Joint Limit Threshold & -$200.0 \cdot \Delta t$ \\
Joint Torque Threshold & -$200.0 \cdot \Delta t$ \\
Joint Velocity Threshold & -$10.0 \cdot \Delta t$ \\
\bottomrule
\end{tabular} \vspace{-5pt}
\end{table}


\begin{table}[t]
\centering
\caption{\textbf{Environment and embodiment randomization scheme in policy training.} 
}
\label{tab:randomization_ranges}
\begin{tabular}{lc}
\toprule
\textbf{Randomization Term} & \textbf{Range} \\
\midrule
Joint Limit Scale & [0, 0.5] \\
Trunk Mass Offset & [-3.0 kg, +7.0 kg] \\
Motor Strength & [50\%, 150\%] \\
PD Control Gains & [50\%, 150\%] \\
Ground Friction & [0.05, 2.0] \\
External Perturbation & $\pm$ 1.0 m/s \\
Gravity Perturbation & $\pm$ 1.0 m/s$^2$ \\
Initial Orientation & Roll/Pitch: $\pm$ 11.25$^\circ$; Yaw: $\pm$ 180$^\circ$ \\
\bottomrule
\end{tabular}\vspace{-10pt}
\end{table}

\subsection{Cross-Embodiment Policy Learning}

We train the locomotion policy with Proximal Policy Optimization (PPO) \cite{schulman2017proximal} in the IsaacLab simulator \cite{mittal2025isaac}. Following~\cite{bohlinger2024onepolicy,ai2025towards}, the policy observes the embodiment description: joint description (axis position and orientation, torque limits, velocity limits, PD gains, nominal position, and range limits), body description (trunk mass only), joint state (position and velocity), and body state (trunk angular velocity, command velocity, and projected gravity). 
The policy predicts joint target positions relative to a nominal standing position, which are tracked by a PD controller with $K_p=20$ and $K_d = 0.5$ \cite{rudin2022learning}. \tabref{tab:rl_reward} summarizes key reward hyperparameters. 

\textbf{Environment and embodiment randomization.}
We use a performance-based curriculum that gradually expands the randomization ranges during training~\cite{bohlinger2025multi, ai2025towards}. This stabilizes learning by exposing the policy to increasingly challenging environments and embodiment variations over time. The final ranges are listed in Table~\ref{tab:randomization_ranges}.


\subsection{Adaptation Module Training}
\textbf{Model implementation details. }
The adaptation module embeds per-joint histories (8-dim) into a feature representation of dimension $F=128$. A per-joint LSTM encodes temporal dependencies, followed by a joint-level Transformer with self-attention for cross-joint interactions~\cite{ai2025towards}. The prediction is generated from the decoder at the last layer, which consists of two MLP heads: a per-joint head for joint-level description and a global head for body-level description.

\textbf{Data collection. }
We generate multiple modified embodiments by sampling discrete joint limit scale and trunk mass offset within predefined ranges. We then collect trajectories for each embodiment. 
In total, we generate 36 distinct embodiment configurations evenly spanning the randomization ranges and collect 4096 trajectories with 1000 timesteps each, totaling 147 million timesteps.
We split the dataset into training and validation sets at a ratio of 8:2.

\begin{figure*}[t]
    \centering
    \includegraphics[width=\textwidth]{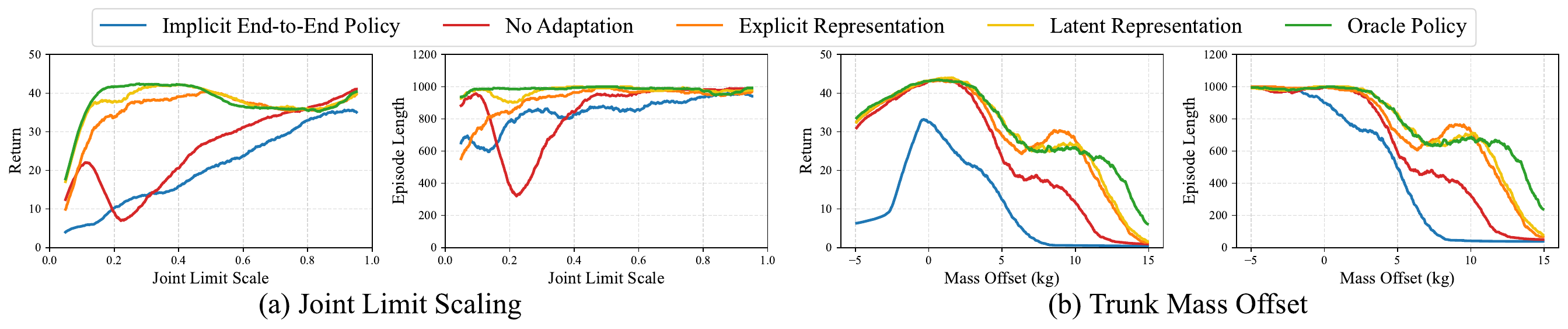}
    \caption{\textbf{Mean episode return and episode length versus joint limit scaling and trunk mass offset in simulation.} Raw values; max 1000 steps = 20 s. We evaluate under (a) varying joint limit scales and (b) varying trunk mass offset.}
    \label{fig:q2_sweep}  \vspace{-10pt}
\end{figure*}

\begin{figure}[t]
    \centering
    \includegraphics[width=\linewidth]{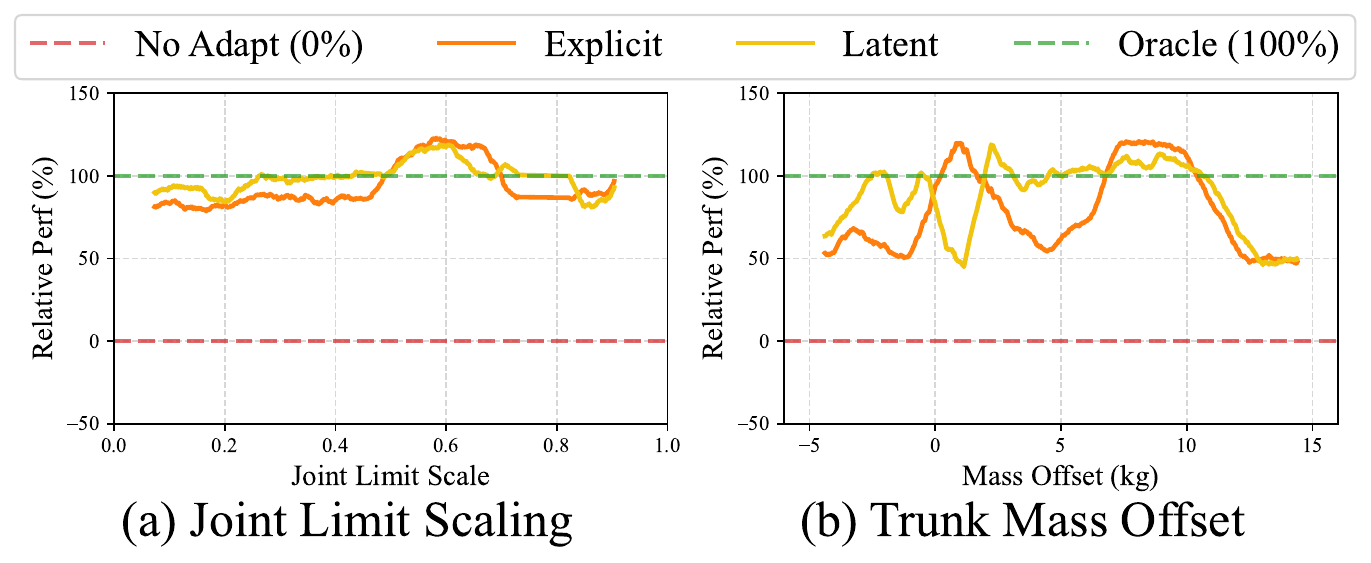}
    \caption{\textbf{Relative performance score $S_{\text{rel}}$ of Explicit and Latent Representations normalized between No Adaptation (0\%) and Oracle Policy (100\%).} We evaluate under (a) varying joint limit scales and (b) varying trunk mass offset.}
    \label{fig:q2_relative} \vspace{-10pt}
\end{figure}

\textbf{Training. }
The adaptation module $f$ ($f_{\text{exp}}$ or $f_{\text{lat}}$) is trained separately via supervised learning on the offline trajectories, minimizing \eqnref{eqn:adapt_loss}. 

\section{Experimental Results} \label{sec:exp_results}

\begin{table}[t]
\centering
\caption{\textbf{Error rate of embodiment parameter prediction on offline validation set.}}
\label{tab:q1_architecture}
\begin{tabular}{lc}
\toprule
\textbf{Model} & \textbf{RMSE} \\
\midrule
MLP & $2.53 \times 10^{-2}$ \\
GRU~\cite{chung2014empirical}          & $2.28 \times 10^{-2}$ \\
\textbf{LSTM~\cite{schmidhuber1997long}}         & $\mathbf{2.02 \times 10^{-2}}$ \\
Transformer~\cite{vaswani2017attention}  & $2.39 \times 10^{-2}$ \\
\bottomrule
\end{tabular} \label{tab:validation}
\vspace{-10pt}
\end{table}

We evaluate the proposed online embodiment adaptation framework through the following questions:
\begin{itemize}[]
    \item[\textbf{Q1.}] Does the adaptation module accurately identify embodiment parameters?
    \item[\textbf{Q2.}] Does online adaptation improve robustness under embodiment uncertainty in simulation?
    \item[\textbf{Q3.}] Does the proposed approach lead to improved performance in the real world?
\end{itemize}

\begin{figure}[t]
    \centering
    \includegraphics[width=1.0\linewidth]{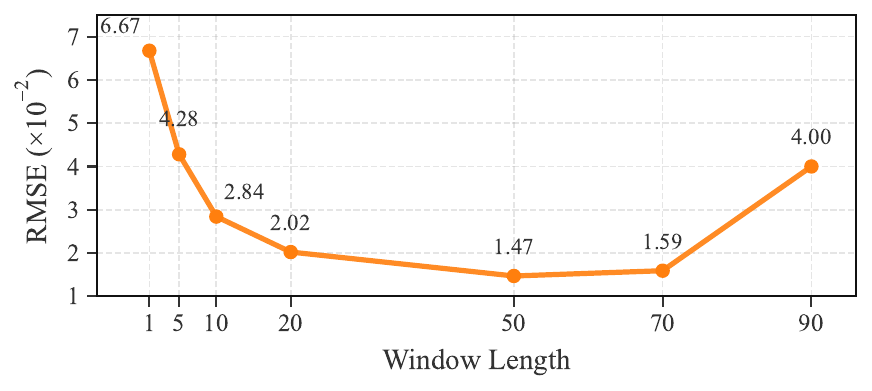}
    \caption{\textbf{Identification RMSE versus history window length on the validation set.}}
    \label{fig:q1_window} \vspace{-10pt}
\end{figure}

\begin{figure}[t]
    \centering
    \includegraphics[width=1.0\linewidth]{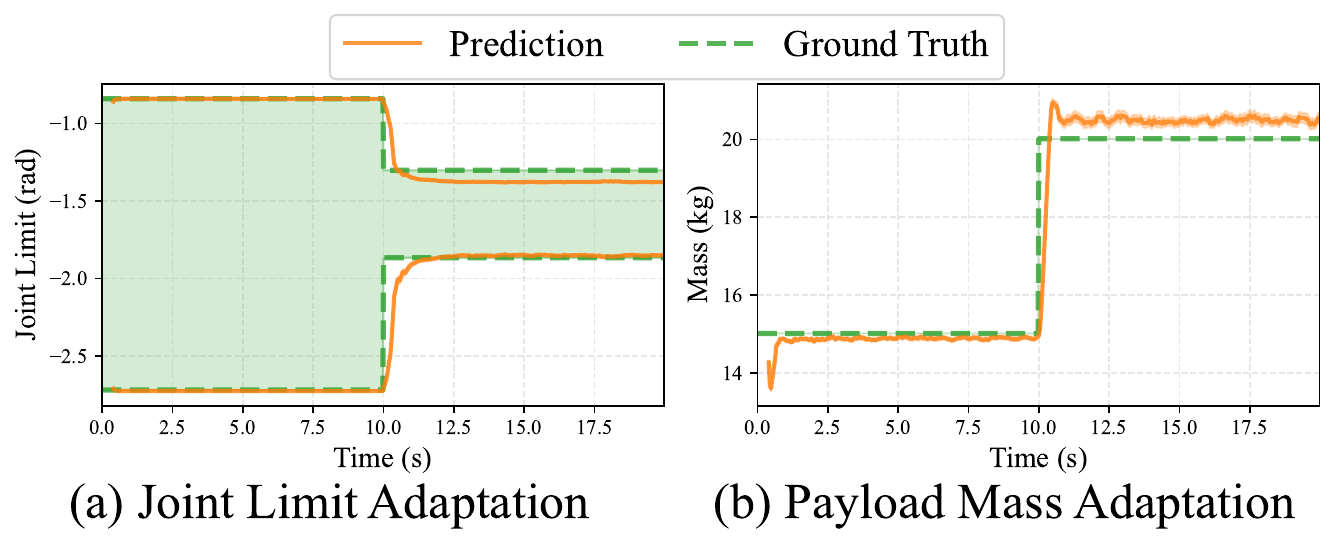}
    \caption{\textbf{Qualitative results on embodiment parameter prediction under 2 abrupt changes: (a) joint limit change; (b) payload mass change.} Our adaptation module is able to identify changing parameters within 0.4 seconds from interaction history.}
    \label{fig:q1_joint_limit_mass_step} \vspace{-10pt}
\end{figure}

\subsection{Offline Evaluation of Embodiment Identification}
\label{sec:q1}

We first evaluate whether the adaptation module can accurately identify embodiment parameters from interactions. We evaluate different design choices by measuring the prediction error on offline datasets. 

\textbf{Architectural comparison.}
We compare different architectures for aggregating history actions and observations for embodiment parameter prediction (\tabref{tab:validation}). LSTM achieves the best performance and is used for the rest of the paper. 

\textbf{History length. } 
Fig.~\ref{fig:q1_window} plots identification loss versus window length. Increasing the window improves accuracy from short to moderate horizons, peaking around 50 timesteps. Beyond that, further increases yield diminishing returns or degradation. We select 20 timesteps as a practical trade-off between accuracy and efficiency.



\textbf{Qualitative results. }
We evaluate under abrupt embodiment changes, where parameters are modified instantaneously during an episode. Our adaptation module (LSTM encoder, 20-timestep window) predicts updated parameters at each timestep.
Fig.~\ref{fig:q1_joint_limit_mass_step} visualizes predictions when joint limits are modified, or payload is added during an episode. Predicted parameters converge to ground truth within 0.4\,s, demonstrating rapid and accurate online adaptation.

\subsection{Closed-Loop Evaluation in Simulation}
\label{sec:q2}

We evaluate the robustness of locomotion under controlled embodiment changes. 
Joint limits are scaled relative to the original per-joint limits. Trunk mass is swept as a continuous payload offset beyond the discrete masses used during training, testing, interpolation, and mild extrapolation.

We compare five models:
(i) \textbf{Implicit End-to-End Policy}: a policy that maps interaction history $h_t$ to actions without estimating embodiment parameters \cite{liu2025locoformer},
(ii) \textbf{No Adaptation}: cross-embodiment base policy with the fixed embodiment description of a canonical Go2 robot,
(iii) \textbf{Explicit Representation}: the base policy conditioned on explicitly predicted embodiment parameters $\hat{\phi}_t$,
(iv) \textbf{Latent Representation}: the base policy conditioned on predicted latent embodiment representations $\hat{\xi}_t$,
and (v) \textbf{Oracle Policy}: the base policy conditioned on the ground-truth embodiment description.

\begin{figure*}[t]
    \centering
    \includegraphics[width=1.\textwidth]{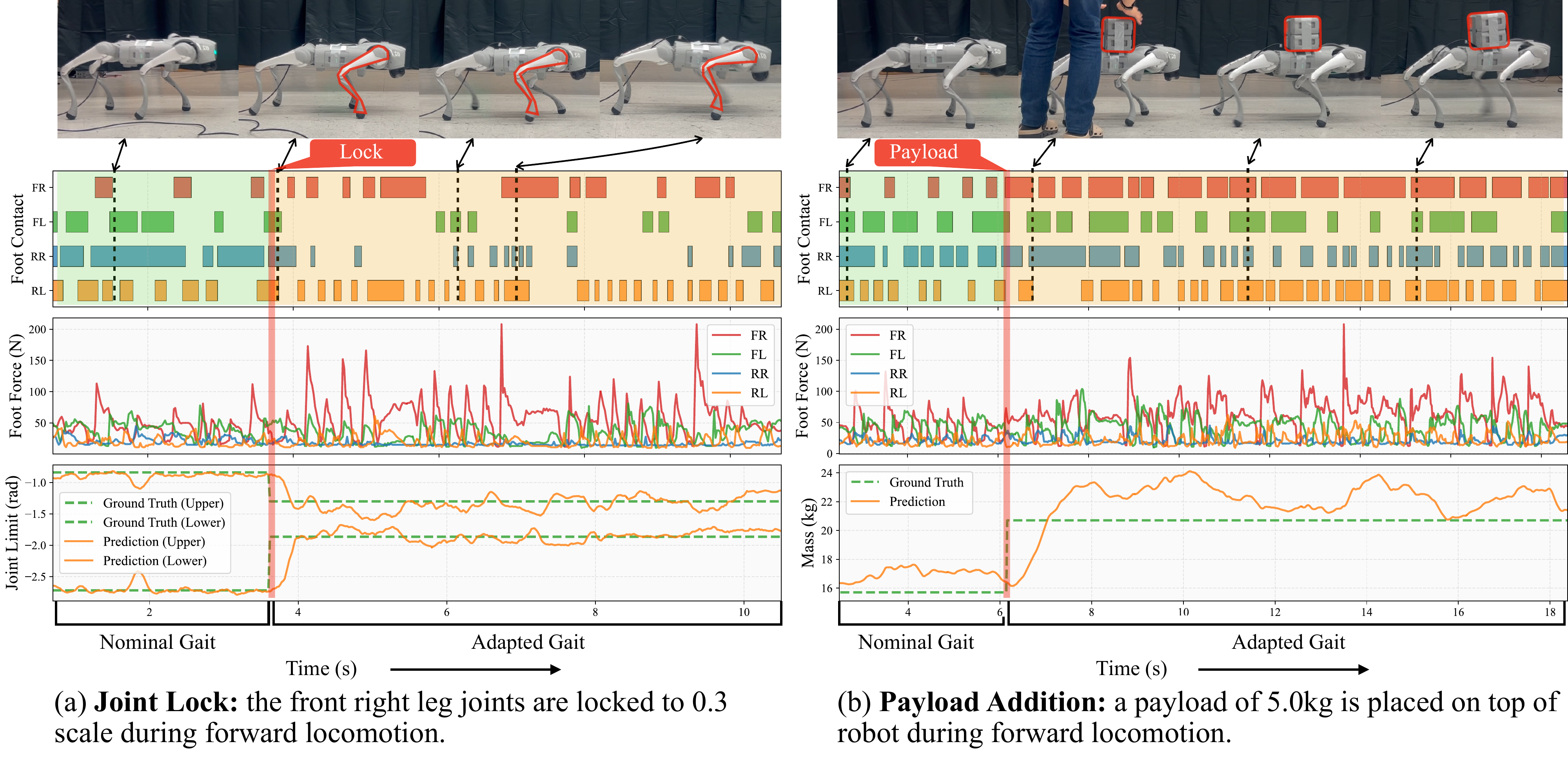}
    \caption{\textbf{Qualitative results for online adaptation under embodiment changes during deployment.} Our adaptation module identifies embodiment parameters within half a second, and the policy conditioned on the inferred parameters adjusts the gait to account for the changed embodiment. \subfig{(a)} When the front-right leg is locked, the robot switches to an asymmetric gait to maintain balance. \subfig{(b)} When the payload is added, the robot shifts to a more load-bearing gait, with longer stance phases and increased foot forces to support the additional mass while maintaining forward locomotion.
    }
    \label{fig:q3_online} 
\end{figure*}

\begin{table*}[t]
\centering
\caption{\textbf{Quantitative metrics over real-world trials under static embodiment uncertainty.} 
Each condition was evaluated for 10 independent trials. All metrics are normalized within each column (higher is better). Success rate, TTF (Time to Failure), and distance are scaled relative to the best observed value; heading deviation is inverted via cosine so that smaller errors yield larger normalized values. 
A score of 1.0 corresponds to a success rate of 80\%, TTF of 13~s, distance of 2.4~m (limited by manual termination), and heading deviation of $3^\circ$.
}
\label{tab:q3_static}
\begin{tabular}{lccccc}
\toprule
\textbf{Condition} & \textbf{Method} & \textbf{Success (norm.)} & \textbf{TTF (norm.)} & \textbf{Distance (norm.)} & \textbf{Deviation (norm.)} \\
\midrule
\textbf{FR Joint Lock 0.1}
& No Adaptation & 0.000 & 0.154 & 0.125 & 0.174 \\
& \textbf{Ours} & \textbf{0.375} & \textbf{0.308} & \textbf{0.417} & \textbf{0.259} \\
\midrule
\textbf{FR Joint Lock 0.3}
& No Adaptation & 0.250 & 0.385 & 0.417 & 0.643 \\
& \textbf{Ours} & \textbf{1.000} & \textbf{1.000} & \textbf{1.000} & \textbf{0.866} \\
\midrule
\textbf{Payload Addition 2.5kg}
& No Adaptation & 0.375 & 0.462 & 0.542 & 0.998 \\
& \textbf{Ours} & \textbf{0.875} & \textbf{0.923} & \textbf{0.917} & \textbf{0.999} \\
\midrule
\textbf{Payload Addition 5.0kg}
& No Adaptation & 0.000 & 0.308 & 0.333 & 0.985 \\
& \textbf{Ours} & \textbf{0.625} & \textbf{0.769} & \textbf{0.750} & \textbf{0.993} \\
\bottomrule
\end{tabular}
\end{table*}


\begin{figure}[t]
    \centering
    \includegraphics[width=\linewidth]{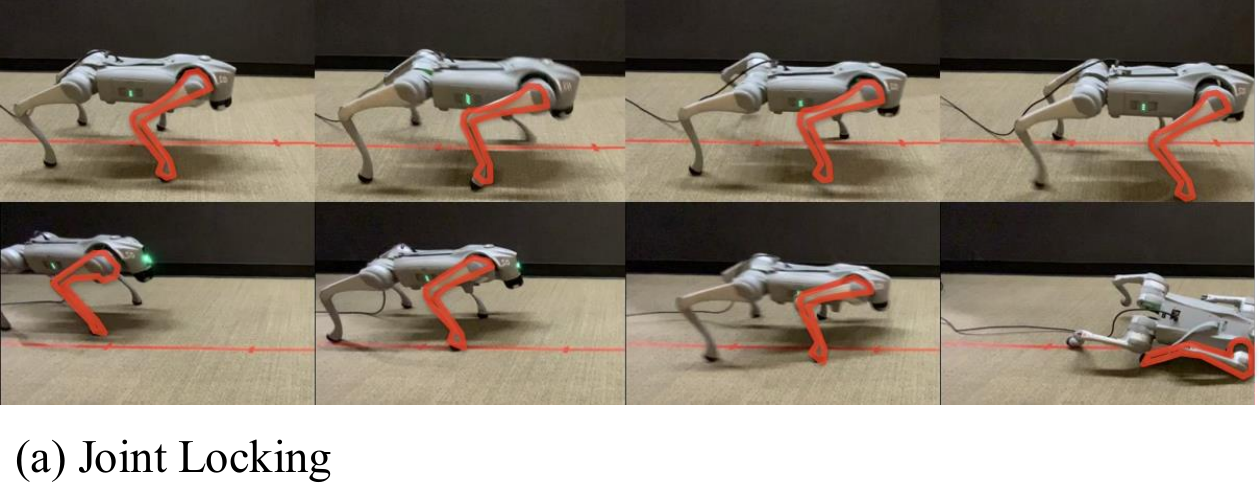}
    \vspace{-0.5em} 
    
    \includegraphics[width=\linewidth]{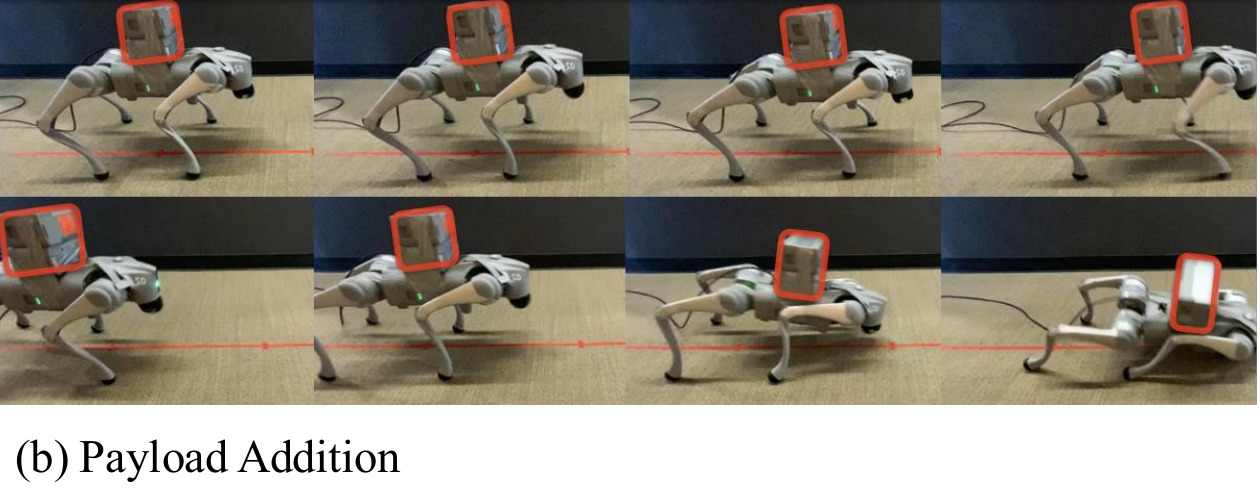}
    
    \caption{\textbf{Gait comparison between our online adaptation (Explicit Representation) method and the No Adaptation baseline under embodiment changes.}
    We evaluate two embodiment changes: (a) restricting the front-right leg joint limits to 0.3 of their nominal range and (b) adding a 5.0\,kg payload to the trunk. The top row in each subfigure shows our method, while the bottom row shows the No Adaptation baseline. In both cases, online adaptation enables stable locomotion, whereas the baseline fails catastrophically.
    }
    \label{fig:q3_static_gait}
\end{figure}

Fig.~\ref{fig:q2_sweep} shows that the No Adaptation degrades rapidly as joint limits and mass offset deviate from the nominal values ($1.0$ and $0.0$ respectively).
The Oracle Policy provides an upper bound. Adaptation using both explicit and latent embodiment representations improves robustness across a wide range of embodiment variations, approaching the Oracle Policy.
The Implicit End-to-End Policy generally performs worse than both proposed adaptation variants across non-nominal embodiment settings, highlighting the effectiveness of explicitly estimating embodiment parameters.

We normalize performance relative to the (ii) No Adaptation and (v) Oracle Policy as lower and upper bounds.
\begin{equation}
S_{\text{rel}} = 
\frac{R - R_{\text{blind}}}{R_{\text{oracle}} - R_{\text{blind}}} \times 100\% ,
\end{equation}
where $R$ denotes the average episode return of a method under the same embodiment setting.
By construction, $S_{\text{rel}}{=}0\%$ and $S_{\text{rel}}{=}100\%$ match the No Adaptation (canonical Go2) and Oracle Policy (ground-truth), respectively.
This normalization quantifies how much of the ``oracle gap'' is closed and avoids over-interpreting absolute returns that vary across settings.
Fig.~\ref{fig:q2_relative} reports $S_{\text{rel}}$ across the joint-limit and trunk-mass sweeps.
Both adaptation methods achieve consistently high $S_{\text{rel}}$, indicating robustness under embodiment uncertainty.

\subsection{Real-World Deployment}
\label{sec:q3}

We validate the explicit embodiment representation in the real world. Given that the implicit end-to-end baseline underperforms both proposed adaptation variants across the simulation sweeps, we focus the hardware evaluation on our framework. Between the explicit and latent variants, we select the explicit representation because its physical parameter estimates are interpretable and directly inspectable during real-world trials.
We demonstrate our method under two conditions: scaling the three front-right leg joint limits and adding payload to the trunk.
Payload is added by placing one or two battery packs (2.5\,kg each) on the trunk.
A trial fails if the robot falls or any joint velocity exceeds 15\,rad/s.

\textbf{Static embodiment. }
We first evaluate robustness when embodiment uncertainty is introduced at episode start.
Tab.~\ref{tab:q3_static} reports metrics averaged over 10 trials; Fig.~\ref{fig:q3_static_gait} shows a representative gait comparison. 
Explicit Representation substantially improves robustness over the No Adaptation across all conditions.
In challenging settings (joint limits locked to 0.1, 5.0\,kg payload), the policy without adaptation fails early while ours maintains stable locomotion longer and travels farther with less drift.
Despite occasional failures in extreme settings, these results show that online adaptation compensates for substantial embodiment uncertainty.

\textbf{Dynamic embodiment changes. }
We evaluate abrupt changes applied mid-episode: (i) sudden reduction of front-right leg joint ranges (scaling to 0.3) and (ii) instantaneous 5.0\,kg payload addition during locomotion.
Fig.~\ref{fig:q3_online} shows representative trials. As shown previously in Fig.~\ref{fig:q3_static_gait}, the base policy typically destabilizes shortly after the change and falls.
In contrast, our method updates the parameters. Predictions converge within approximately 20 timesteps ($\approx 0.4$s at 50\,Hz), and the policy transitions to a gait adapted to the new constraints.
For joint limits, the robot adopts an asymmetric gait: the FR leg acts as a crutch in continuous contact while the FL leg swings for forward motion, and stance/swing is redistributed across the remaining legs to keep balance.
For payload, the policy lowers the trunk and increases ground contact duration of all four legs to support the load, while maintaining a symmetric gait.

%% file: text/060_conclusion.tex
\section{Discussion} \label{sec:conclusion}


We presented an online embodiment adaptation framework for quadrupedal locomotion. The framework infers an explicit embodiment description from short interaction histories and conditions a cross-embodiment policy on the inferred parameters. On the real-world Unitree Go2, adaptation converges within half a second, sustaining stable locomotion under a fully locked leg and a 5 kg payload where non-adaptive methods fail.

Beyond these results, the framework relaxes a common assumption in embodiment-conditioned policy deployment: that embodiment information is known and static. Our experiments focus on joint-range and trunk-mass changes within a single robot platform, providing a foundation for extending online identification to broader variations in morphology and actuation.


%% file: text/070_appendix.tex
\clearpage